%% file: main.tex
\RequirePackage{pdf14}

\documentclass[sigconf]{acmart}
\usepackage{url}
\usepackage{multirow}
\usepackage{subcaption}
\usepackage{array}
\usepackage{colortbl}
\usepackage{xcolor}

\usepackage{framed}
\usepackage{mdwlist}
\usepackage{latexsym}
\usepackage{colortbl}
\usepackage{xcolor}
\usepackage{nicefrac}
\usepackage{booktabs}
\usepackage{amsfonts}
\usepackage[T1]{fontenc}
\usepackage{bold-extra}
\usepackage{amsmath}
\usepackage{bm}
\usepackage{graphicx}
\usepackage{mathtools}
\usepackage{microtype}
\usepackage{multirow}
\usepackage{multicol}
\usepackage{latexsym,comment}
\usepackage[normalem]{ulem}
\usepackage{hyphenat}

\AtBeginDocument{%
  \providecommand\BibTeX{{%
    \normalfont B\kern-0.5em{\scshape i\kern-0.25em b}\kern-0.8em\TeX}}}
\makeatother

\newif\ifcomment\commentfalse

\newcommand{\camelabr}[2]{{\small #1}{\textsc{#2}}}
\newcommand{\abrcamel}[2]{{\textsc #1}{\small{#2}}}

\newcommand{\gem}[1]{\mbox{\textsc{gem}}}

\newcommand{\explain}[2]{\underbrace{#2}_{\mbox{\footnotesize{#1}}}}

\newcommand{\g}{\, | \,}

\newcommand{\name}{\abr{delft}}

\newcommand{\bertet}{\abr{bert-e}{\small ntity}}
\newcommand{\bertsent}{\abr{bert-s}{\small ent}}
\newcommand{\memnn}{\abr{bert-m}{\small em}\abr{nn}}
\newcommand{\kgqa}{\abr{kgqa}}
\newcommand{\quest}{\abr{quest}}

\newcommand{\abr}[1]{\textsc{#1}}
\newcommand{\qb}[0]{Quizbowl}

\newcommand{\triviaqa}{\camelabr{Trivia}{qa}}

\newcommand{\qblink}{\abrcamel{qb}{Link}}
\newcommand{\qanta}{\textsc{qanta}}

\newcommand{\squad}{\textsc{sq}{\small u}\textsc{ad}}

\newcommand{\jeopardy}{\textit{Jeopardy!}}

\newcommand{\glove}{\abr{gl}{\small o}\abr{ve}}
\newcommand{\openie}{\abr{o}{\small pen}\abr{ie}}
\newcommand{\bert}{\abr{bert}}
\newcommand{\drqa}{\abr{d}{\small r}\abr{qa}}
\newcommand{\docqa}{\abr{d}{\small oc}\abr{qa}}

\newcommand{\sigmoid}{Sigmoid}

\newcommand{\ffn}[1]{\mbox{\abr{ffn}}\left(#1\right)}
\newcommand{\rnn}[1]{\mbox{\abr{rnn}}\left(#1\right)}

\newcommand{\softmax}[1]{\mbox{softmax}\left(#1\right)}
\newcommand{\avg}[2]{\mbox{Avg}_{#2}\left(#1\right)}

\newcommand{\question}[1]{\texttt{#1}}
\newcommand{\edge}[1]{``#1''}
\newcommand{\candidate}[1]{\underline{#1}}

\newcommand{\leftnode}{Question Entity Node}
\newcommand{\rightnode}{Candidate Entity Node}
\newcommand{\tweennode}{Evidence Edge}
\newcommand{\kg}{\abr{kg}}
\newcommand{\dbpedia}{\abr{db}{\small pedia}}

\newcolumntype{g}{>{\columncolor{lightgray}}r}

\begin{document}

\setcopyright{acmcopyright}

\copyrightyear{2020}
\acmYear{2020} 
\acmConference[WWW '20]{Proceedings of The Web Conference 2020}{April 20--24, 2020}{Taipei, Taiwan} 
\acmBooktitle{Proceedings of The Web Conference 2020 (WWW '20), April 20--24, 2020, Taipei, Taiwan}
\acmPrice{}
\acmDOI{10.1145/3366423.3380197}
\acmISBN{978-1-4503-7023-3/20/04}


\title{Complex Factoid Question Answering with a Free-Text Knowledge Graph}

\author{Chen Zhao}
\affiliation{%
  Computer Science, \abr{umiacs} \\
  \institution{University of Maryland, College Park}
}
\email{chenz@cs.umd.edu}

\author{Chenyan Xiong}
\affiliation{%
  \institution{Microsoft AI \& Research}
}
\email{cxiong@microsoft.com}

\author{Xin Qian}
\affiliation{%
  iSchool \\
  \institution{University of Maryland, College Park}
}
\email{xinq@terpmail.umd.edu}

\author{Jordan Boyd-Graber$^{\dagger}$}
\affiliation{%
  \thanks{$^{\dagger}$Now at Google Research Z\"urich}
    Computer Science, iSchool, \abr{umiacs}, \abr{lsc} \\
  \institution{University of Maryland, College Park}
}
\email{jbg@umiacs.umd.edu}

\input{2020_www_delft/sections/00-abstract} 

\maketitle

\input{2020_www_delft/sections/10-intro}	
\input{2020_www_delft/sections/20-justification}

\input{2020_www_delft/sections/30-evidence}

\input{2020_www_delft/sections/50-model}

\input{2020_www_delft/sections/70-experiment}

\input{2020_www_delft/sections/71-evaluation}

\input{2020_www_delft/sections/60-related}

\input{2020_www_delft/sections/80-conclusions}

\section*{Acknowledgements}

This work was supported by NSF Grants IIS-1748663 (Zhao) and
IIS-1822494 (Boyd-Graber).  Any opinions, findings, conclusions, or
recommendations expressed here are those of the authors and do not
necessarily reflect the view of the sponsor.



\bibliographystyle{style/ACM-Reference-Format}
\normalsize
\newpage

\bibliography{bib/journal-full,bib/chen,bib/jbg}

\end{document}

%% file: 2020_www_delft/sections/00-abstract.tex
\begin{abstract}

We introduce \name{}, a factoid question answering system which
combines the nuance and depth of knowledge graph
question answering approaches with the broader coverage of free-text.
\name{} builds a free-text knowledge graph from Wikipedia, 
with entities as nodes and sentences in which entities co-occur as edges.
For each question, \name{} finds the subgraph linking question entity
nodes to candidates using text sentences as edges, creating a dense
and high coverage semantic graph.
A novel graph neural network reasons over the free-text graph---combining
evidence on the nodes via information along edge
sentences---to select a final
answer.
Experiments on three question answering datasets show \name{} can
answer entity-rich questions better than machine reading based
models, \abr{bert}-based answer ranking and memory networks.
\name{}'s advantage comes from both the high coverage of its free-text
knowledge graph---more than double that of \dbpedia{} relations---and
the novel graph neural network which reasons on the rich but 
noisy free-text evidence.

\end{abstract} 

%% file: 2020_www_delft/sections/10-intro.tex
\section{Introduction}
\label{sec:intro}


Factoid question answering ~\cite[\abr{qa}]{wang2006survey,iyyer2014neural, gardner2019question}, which asks precise facts about entities, 
is a long-standing problem in
natural language processing (\abr{nlp}) and information retrieval (\abr{ir}).
A preponderance of knowledge graphs (\abr{kg})
such as Freebase~\cite{bollacker2008freebase}, \dbpedia{}~\cite{mendes2011dbpedia}, 
\abr{yago}~\cite{suchanek2007yago}, and Wikidata~\cite{vrandevcic2014wikidata}
have been fruitfully applied to factoid \abr{qa} over knowledge
graphs (\abr{kgqa})~\cite[inter alia]{bordes2014question, dong-etal-2015-question}. 
A typical \abr{kgqa} task is to search for entities or
entity attributes from a \abr{kg}.
For example, given the question ``who was the last emperor of China'',
the model identifies emperors of China and then reasons to determine
which emperor is the last and find the answer, \candidate{Puyi}.

\kgqa{} approaches solve the problem by
either parsing questions into executable logical
form~\cite{cai2013large, kwiatkowski2013scaling} or aligning questions
to sub-structures of a knowledge graph~\cite{yao2014information,
  yih2015semantic}.
Whether these approaches work depends on the underlying \abr{kg}: they
are effective when the knowledge graphs have high coverage 
on the relations targeted by the questions, which is often not the case:
%
Obtaining relation data is costly and requires expert knowledge 
 to design the relation schema and type
 systems~\cite{Paulheim2018HowMI}.
For example, Google's Knowledge Vault has 570M entities, fourteen times more
than Freebase, but only has 35K relation types: similar to
Freebase~\cite{knowledgevalut, bollacker2008freebase}.

\begin{figure*}[t]
    \begin{center}
    \includegraphics[width=0.9\linewidth]{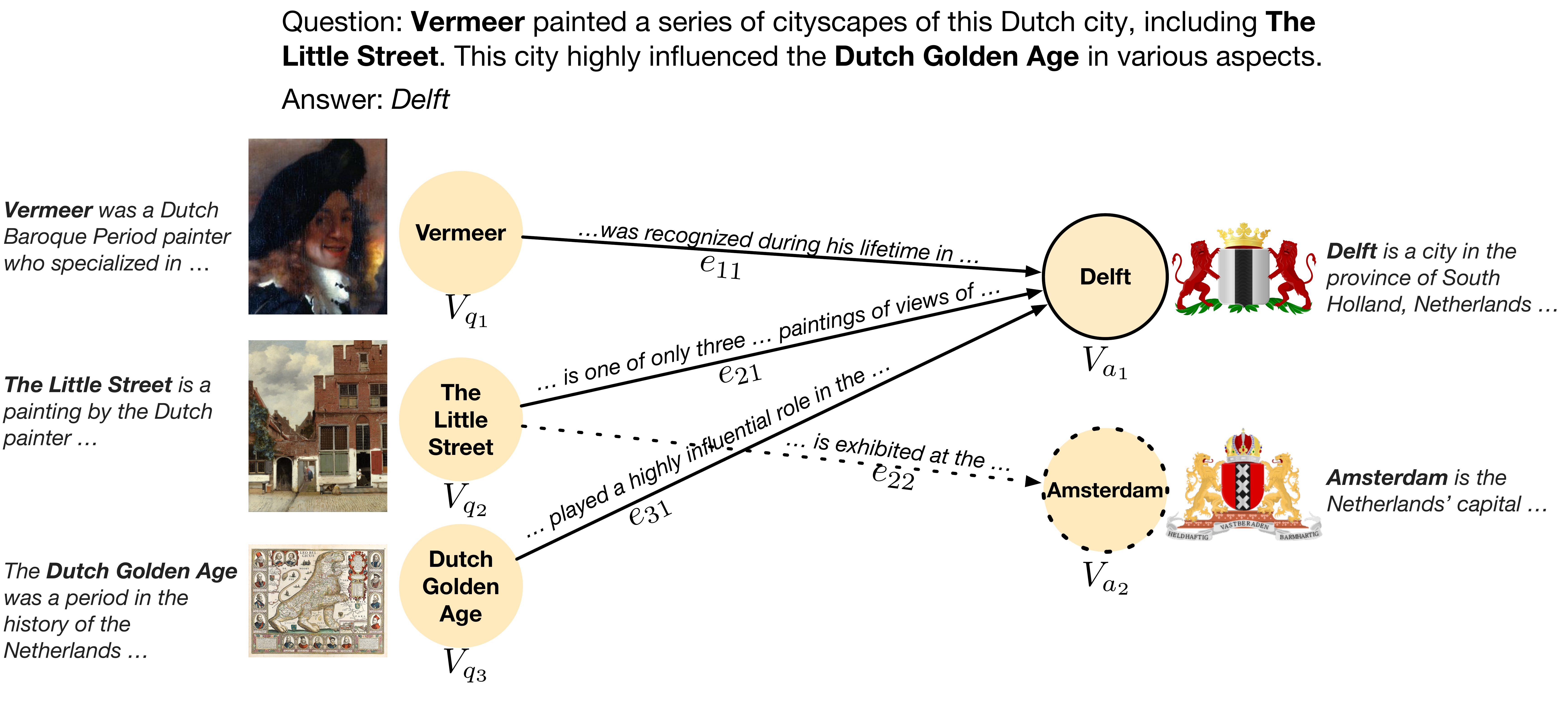}
    \end{center}
    \caption{An example question grounded free-text knowledge graph in
      \name{}. The graph has both question entity nodes (left side)
      and candidate entity nodes (right side), and use natural
      language sentences from Wikipedia for both nodes and edges.}
    \label{fig:graph}
\end{figure*}

In the real world, human knowledge is often tested through
competitions like Quizbowl~\cite{boyd-graber-12} or
\jeopardy{}~\cite{ferruci-10}.
%
In these settings, the question clues are complex, diverse, and
linguistically rich,
most of which are unlikely to be covered by the well-formatted 
closed form relations in knowledge graphs,
and are often phrased obliquely, 
making existing \kgqa{} methods brittle.
Figure~\ref{fig:graph} shows a real world factoid \abr{qa} example.
The relations in the question are complex and defy categorization into
\kg{} relationships.
For example, the relation ``depict''---much less ``paint a cityscape of''---from the first
question sentence ``Vermeer painted a series of cityscapes of this
Dutch city'' is not a frequent \kg{} relation.

One possible solution to sparse coverage is machine
reading~(\abr{mr}), which finds an answer directly from unstructured
text~\cite{chen2018neural}.
Because it extracts answers from paragraphs and documents---pre-given
or retrieved~\cite{rajpurkar2016squad, clark2018simple,
  zhong2019coarse}---\abr{mr} has much broader coverage~\cite{marco}.
However, current \abr{mr} datasets mainly evaluate on reasoning
within a single paragraph~\cite{min2018efficient}; existing
\abr{mr} models focus on extracting from single evidence.
Synthesizing multiple pieces of evidence to answer
complex questions remains an on-going research 
topic~\cite{yang2018hotpotqa, min2019compositional}.

To build a \abr{qa} system that can answer real-world, complex factoid
questions 
using unstructured text, 
we propose \name{}: \textbf{D}eciphering \textbf{E}ntity
\textbf{L}inks from \textbf{F}ree \textbf{T}ext.
\name{} constructs a free-text knowledge graph from Wikipedia, with
entities (Wikipedia page titles) as nodes, and---instead of depending
on pre-defined relations---\name{} uses free-text sentences as edges.
This subgraph are the nodes relevant to the question, grounded to text
by connecting question entities to candidate answer entities with
extracted free-text sentences as evidence edges.
The constructed graph supports complex modeling over
its graph structures and also benefits from the high coverage of
free-text evidence.

Unlike existing knowledge graphs, in which the relations between
entities are well-defined, \name{} leverages informative, diverse, but noisy
relations using a novel graph neural network (\abr{gnn}).
Each edge in our graph is associated with multiple sentences that give
explain how entities interact with each other. \name{} uses the \abr{gnn} to distinguish the 
useful evidence information from unrelated text and aggregates
multiple evidence from both text and the graph structure to answer the question.
%
%

We evaluate \name{} on three question answering
datasets: \qblink{},  \qanta{}, and \triviaqa{}.
\name{} outperforms machine reading based model~\cite{chen2017reading},
\abr{bert}-based answer ranking~\cite{devlin2018bert}, and a \abr{bert}-based memory
network approach~\cite{weston2014memory} that uses the same evidence
but no graph structure, with significant margins.
Its accuracy improves on more complex questions and when dense
evidence is available, while \abr{mr} cannot take advantage of
additional evidence.

Ablation reveals the importance of each model component: node
representation, edge representation, and evidence combination.
Our graph visualization and case studies further illustrate that our
model could aggregate multiple pieces of evidence to make a more
reliable prediction.\footnote{The code, data, full free-text knowledge graph, and question grounded subgraph
is at \url{delft.qanta.org}. }

%% file: 2020_www_delft/sections/20-justification.tex
\section{Task Definition}
\label{sec:justification}

The problem we consider is general \textit{factoid question
  answering}~\cite[inter alia]{iyyer2014neural, berant2013semantic, bordes2015large, 
  cai2013large}, in which the
system answers natural language questions using facts, i.e., entities
or attributes from knowledge graphs.
Factoid \abr{qa} is both widely studied in academia and a practical
tool used by commercial search engines and conversational assistants.
However, there is a discrepancy between the widely studied factoid
\abr{qa} academic benchmarks and  real applications:
the benchmarks are often designed for the existing relations in
knowledge graphs~\cite{berant2013semantic,bordes2015large}, while in reality,
minimal \abr{kg} coverage preculdes broader deployment.

For example, WebQuestions~\cite{berant2013semantic} are
written by crowdworkers targeting a triple in Freebase; \abr{l}{\small
  c}-quad~\cite{dubey2019lc} targets Wikidata and \dbpedia{}: the
questions are guaranteed to be covered by the \abr{kg} relations.
However, although the entities in your favorite
\abr{kg} are rich, the recall of closed-form relations is often
limited.
For example, in Figure~\ref{fig:graph}, the answer \underline{Delft}
is an entity in Wikipedia and Freebase, but the relation ``painting
the view of'' appears in neither DBpedia nor Freebase and is unlikely
to be covered in other knowledge graphs.
The relationships between complicated, multi-faceted entities defy
trite categorization in closed-form tuples; depending on a knowledge
graph to provide all the possible relationships limits the potential
of \abr{kgqa} systems.

We focus on a more realistic setting: open domain factoid
question answering, whose questions are designed to test \emph{human}
knowledge and include ineffable links between concepts~\cite{jennings-06}.
The datasets in our evaluation are solvable by humans using knowledge
about real world entities; \name{} should also be able to find the
answer entity without relying on explicit closed form relations.

Figure~\ref{fig:graph} shows an example question. 
The question has multiple \leftnode{}s (left side of graph), 
and links the question entities to \rightnode{}s (right). 
Intuitively, to find the answer, the system needs to first extract
multiple clues from different pieces of question.
Our proposed \name{} constructs a high coverage Free-Text Knowledge
Graph by harvesting natural language sentences in the corpus as graph
edges and grounds each question into its related subgraph
(Section~\ref{sec:evidence}).
Then, to model over the fruitful but noisy graph, it uses a \abr{gnn}
to distinguish useful evidence from noise and then aggregates them to
make the prediction (Section~\ref{sec:model}).

%% file: 2020_www_delft/sections/30-evidence.tex
\section{Free-Text Graph Construction}
\label{sec:evidence}

Answering real world factoid questions with current knowledge
graphs falters when \kg{} relations lack coverage; we instead
use a free-text \kg{} to resolve the coverage
issue. One attempt to incorporate free-text corpus is to use Open
Information Extraction~\cite[\openie]{Lu:2019:ACQ} to extract
relations from natural language sentences as graph edges.
However \openie{} approaches
favor precision over recall, and heavily rely on well defined semantics, falling prey to the same
narrow scope that makes traditional \abr{kg} approaches powerful.
%
Instead, we build the knowledge graph directly from free-text corpus,
represent indirect usings sentences which contain the two entities
(endpoints of the edge) as indirect relations.
We leave the \abr{qa} model (which only needs to output an answer
entity) to figure out \emph{which} sentences contain the information
to answer the question, eliminating the intermediate information
extraction step.

This section first discusses the \textit{construction} of a
\emph{free-text} knowledge graph and then \textit{grounding} questions
to it.

\subsection{Graph Construction}

The free-text knowledge graph uses the same nodes $V$ as
existing knowledge graphs: entities and their attributes, which can be
directly inherited from existing knowledge graphs.
The \tweennode{}s~$E$, instead of closed-form relations, are harvested natural
language sentences from a corpus.

\paragraph{Entity Nodes}

The graph inherits Wikipedia entities as the nodes $V$ (of course,
entities from other corpora could be used).
To represent the node, we use the first sentence of the corresponding
document as its \textit{node gloss}.

\paragraph{Free-Text Edges}

The \tweennode{}s between nodes are sentences that pairs of entities
in Wikipedia co-occur (again, other corpora such as
ClueWeb~\cite{callan2009clueweb09} could be used).
For specificity, let us find the edges that could link entities $a$ and $b$.
First, we need to know where entities appear.
Both $a$ and $b$ have their own Wikipedia pages---but that
does not give us enough information to find where the entities appear
\emph{together}.
TagMe~\cite{ferragina2010tagme} finds entities mentioned in free text;
we apply it to Wikipedia pages.
Given the entity linker's output, we collect the following sentences
as potential edges: (1) sentences in $a$'s Wikipedia page that mention
$b$, (2) sentences in $b$'s page that mention $a$, and (3) sentences
(anywhere) that mention both $a$ and $b$.

\subsection{Question Grounding}

Now that we have described the general components of our graph, we
next \textit{ground} a natural language question to a
subgraph of the \abr{kg}.  We then find an answer candidate in
this subgraph.

Specifically, for each question, we ground the full free-text
knowledge graph into a question-related subgraph
(Figure~\ref{fig:graph}): a bipartite graph with \emph{\leftnode{}s}
(e.g., \question{Dutch Golden Age}) on the left joined to
\emph{\rightnode{}s} on the right (e.g., \candidate{Delft}) via
\emph{\tweennode{}} (e.g., \edge{\candidate{Delft} played a highly
  influential role in the \question{Dutch Golden Age}}).

\paragraph{\leftnode{}s}

\name{} starts with identifying the entities in
question $\mathcal{X}_q$ as the \leftnode{}s $V_q=\{v_{i} \g v_i \in
\mathcal{X}_q\}$.
In Figure~\ref{fig:graph}, for instance, \question{Vermeer},
\question{The Little Street} and \question{Dutch Golden Age} appear in
the question; these \leftnode{}s populate the left side of
\name{}'s grounded graph. We use TagMe to identify question entities.


\paragraph{\rightnode{}s}

Next, our goal is to find \rightnode{}s.
In our free-text \abr{kg}, \rightnode{}s are the entities with 
 connections to the entities related to the question.
For example, \candidate{Delft} occurs in the Wikipedia page associated
with \leftnode{} \question{Vermeer} and thus becomes
a \rightnode{}.
The goal of this step is to build a set likely to
contain---has high coverage of---the answer entity.

We populate the \rightnode{} $V_a$ based 
on the following two approaches.
First, we \textit{link} entities contained in the Wikipedia \emph{pages} of
the \leftnode{}s to generate \rightnode{}s.
%
%
To improve \rightnode{} recall, we next \textit{retrieve} entities:
after presenting the question text as a query to an \abr{ir} engine
(ElasticSearch~\cite{Gormley:2015:EDG:2904394}), the entities in the
top retrieved Wikipedia pages also become \rightnode{}s.

These two approaches reflect different ways entities are mentioned
in free text.
Direct mentions discovered by entity linking tools are straightforward
and often correspond to the clear \emph{semantic} information encoded in
Wikipedia (``In 1607, Khurram became engaged to Arjumand Banu Begum,
who is also known as Mumtaz Mahal'').
However, sometimes relationships are more \emph{thematic} and are not
mentioned directly; these are captured by \abr{ir} systems
(``She was a recognizable figure in academia, usually wearing a
distinctive cape and carrying a walking-stick'' describes
\underline{Margaret Mead} without named entities).
Together, these two approaches have good recall of the \rightnode{}
set $V_a$ (Table~\ref{tab:coverage}).

\paragraph{\tweennode{}s}

\name{} needs edges as evidence signals to know which \rightnode{} to select.
The edges
connect \leftnode{}s $V_q$ to \rightnode{}s
$V_a$ with natural language.
In Figure~\ref{fig:graph}, the Wikipedia sentence
\edge{Vermeer was recognized during his lifetime in Delft} connects the
\leftnode{} \question{Vermeer} to the \rightnode{}
\candidate{Delft}.
%
Given two entities, we directly find the \tweennode{}s connecting them
from the free-text knowledge graph.

\paragraph{Final Graph}

Formally, for each question, our final graph of \name{} $G =(V, E)$
includes the follows:
Nodes $V$ include \leftnode{}s $V_q$ and \rightnode{}s
$V_a$;
\tweennode{}s $E$ connect the nodes: each edge~$e$ contains Wikipedia
sentence(s) $\mathcal{S}({k}), k= 1,\dots, K$ linking the nodes.

\subsection{Question Graph Pruning}

The graph grounding ensures high coverage of nodes and edges.
However, if the subgraph is \emph{too} big it slows training and inference.
\name{} prunes the graph with a simple filter to remove weak, 
spurious clues and to improve computational efficiency.

\paragraph{Candidate Entity Node Filter}

We treat the \rightnode{}s filtering as a ranking problem: given input
\rightnode{}s and question, score each connected \tweennode{} with a relevance score. 
During inference, the nodes with top-$K$ highest scores are kept (we choose the highest \tweennode{} score as the node relevance score), while
the rest are pruned. 
\name{} fine-tunes \abr{bert} as the filter model.\footnote{For efficiency, We use \abr{tfidf} filtering (top 1000 \tweennode{}s are kept) before \abr{bert}.}
To be specific, for each connected \tweennode{},
we concatenate the question and sentence, along with the \rightnode{}
gloss as the node context into \abr{bert}:
%
\begin{verbatim}
 [CLS] Question [SEP] Edge Sent [SEP] Node Gloss [SEP]
\end{verbatim}
We apply an affine layer and sigmoid 
activation on the last layer's 
[CLS] representation to produce scalar value.

During training, for each question, we use \tweennode{} connected to the answer node as positive
training example, and random sample $10$ negative \tweennode{}s as
negative examples. To keep the \name{}'s \abr{gnn} training efficient,
we keep the top twenty nodes 
 for training set. For more
comprehensive recall, development and test keep the top fifty nodes.
If the answer node is not kept, \name{} cannot answer the question.

\paragraph{Edge Sentence Filter}

Since there is no direct supervision signal for which edge sentences
are useful, we use \abr{tfidf}~\cite{salton1983introduction} to filter
sentences.
For all sentences $\mathcal{S}$ in \tweennode{} $e \in E$, we compute
each sentence's \abr{tfidf} cosine similarity to the question, and choose the
top five sentences for each \tweennode{}.

%% file: 2020_www_delft/sections/50-model.tex
\begin{figure}[h!]

    \begin{center}
    \includegraphics[width=0.86\linewidth]{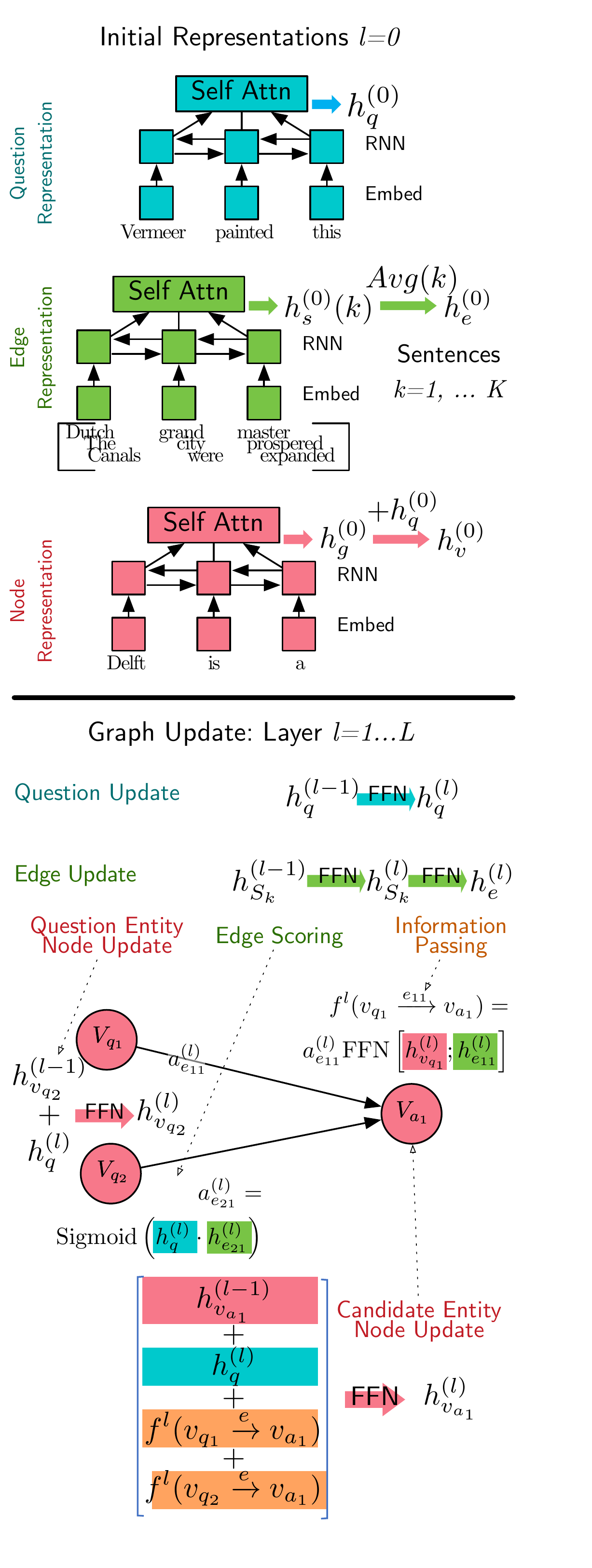}
    \end{center}
    \caption{Model architecture of \name{}'s \abr{gnn}. The left side
      shows the initial representation of the network. The right side
      illustrates the graph update.  }
    \label{fig:model}
\end{figure}

\section{Free-text Graph Modeling}
\label{sec:model}

Given the question $\mathcal{X}_q$ and the grounded free-text graph
$G$ from Section~\ref{sec:evidence}, the goal is to find the correct
answer node from the \rightnode{}s.  Recently several \abr{gnn} based
approaches~\cite{tu-etal-2019-multi, de2018question} answer questions
by modeling the knowledge graph.
%
Unlike previous approaches that represent graph nodes and edges as fixed embeddings, 
\name{} adopts a \abr{gnn} to find the answer using  
a free-text knowledge graph.
We make the following motivating observations:

\paragraph{Graph Connectivity}

The correct \rightnode{} usually has many connections to \leftnode{}s.
For example, in Figure~\ref{fig:graph}, the correct \rightnode{}
\candidate{Delft} connects to all three question \leftnode{}s.
Thus, a correct \rightnode{} should \textit{aggregate} information
from multiple \leftnode{}s.

\paragraph{Edge Relevance}
The \tweennode{}s with sentences ``closer'' to the question
are likely to be more helpful to answering.
For example, in Figure~\ref{fig:graph}, the \leftnode{}
\question{The Little Street} has two edges to \rightnode{}s
\candidate{Delft} and \candidate{Amsterdam}, but the \tweennode{} with sentence
``\question{The Little Street} is one of only three paintings of views of \candidate{Delft}''
is more similar to the question.
The model needs to prioritize \tweennode{}s that are similar to the question.

\paragraph{Node Relevance}

In addition to relevant edges, we also use node
information to focus attention on the right \rightnode{}.
One aspect is ensuring we get the entity type correct (e.g., not
answering a place like \candidate{Uluru} to a ``who'' question like
``who took managing control of the Burra Burra mine in 1850?'').
For both \leftnode{} and \rightnode{}, we use entity gloss sentences
as node features.
For example, the gloss of the candidate entity node says
``\candidate{Delft} is a Dutch city'', which aligns what the question
asks (``this dutch city'').
Similarly, a \leftnode{} whose gloss sentence better matches a
question are more likely to contain the clues for the answer.

\paragraph{Iterative Refinement}

Figure~\ref{fig:model} illustrates the architecture of \name{}'s
\abr{gnn}.
A key component of the \abr{gnn} is that multiple layers ($^{(l)}$)
refine the representations of \rightnode{}s, hopefully focusing on the
correct answer.

The first layer learns the initial representation of the nodes and
edges, using their textual features and the relevance to the question
(Section~\ref{sec:init}).
Each successive layer combines this information, focusing on the
\tweennode{}s and \rightnode{}s that can best answer the question
(Section~\ref{sec:prop}).
The final layer scores the \rightnode{}s to produces an answer
(Section~\ref{sec:ans}).

\subsection{Initial Representations}
\label{sec:init}

We start with representations for questions, \leftnode{}s,
\rightnode{}s, and \tweennode{}s (\abr{gnn} layer~0).  These
representations are combined in subsequent layers.

\paragraph{Question and Node Representation}

Each question $\mathcal{X}_q$ and node gloss (both \rightnode{} and \leftnode{}) $\mathcal{X}_{g}$ 
is a sequence of word tokens $\mathcal{X} = (x_1,\dots, x_n)$.  Individual word embeddings $(\mathbf{E}_{x_1},\dots,
\mathbf{E}_{x_n})$ become a sequence representation~$\mathbf{h}_q^{(0)}$ ($\mathbf{h}_g^{(0)}$) using a 
Recurrent neural network (\abr{rnn}) with a self-attention (\abr{self-attn}) layer:
\begin{equation}
\mathbf{h}_{x_u} = \rnn{\mathbf{E}_{x_1},\dots, \mathbf{E}_{x_n}}
\end{equation}
where $u$ is token position of sequence~$\mathcal{X}$.  A
self-attention layer over the \abr{rnn} hidden layer first weights the hidden states
by a learnable weight vector $\mathbf{w}_x$
\begin{equation}
  a_{x_u} = \softmax{\mathbf{w}_x \cdot \mathbf{h}_{x_u}},
\label{eq:self_attn}
\end{equation}
then a final representation ${h}_x^{(0)}$ is a weighted average of 
all hidden states
\begin{equation}
\mathbf{h}_x^{(0)} = \sum_u a_{x_u}  \mathbf{h}_{x_u} .
\end{equation}

The node representation $\mathbf{h}_{v}^{(0)}$ of each node~$v~\in V$
is a sum of gloss representation $\mathbf{h}^{(0)}_{g}$ 
and question representation $\mathbf{h}_q^{(0)}$.
\begin{equation}
    \mathbf{h}_{v}^{(0)} = \mathbf{h}^{(0)}_{g} + \mathbf{h}_q^{(0)}.
\end{equation}
The representation $\mathbf{h}_{v}^{(0)}$ is applied to both question
entity nodes $\mathbf{h}_{v_q}^{(0)}$ and candidate answer nodes
$\mathbf{h}_{v_a}^{(0)}$.

\paragraph{Edge Representation}

Effective \tweennode{}s in \name{} should point from entities
mentioned in the question to the correct \rightnode{}.
We get the representation $\mathbf{h}_{e}^{(0)}$ of edge $e~\in E$
by first embedding each edge sentence $S(k)$'s tokens $\mathcal{X}_{s} = (s_1, ..., s_n)$
%
into $(\mathbf{E}_{s_1}, ..., \mathbf{E}_{s_n})$, then encoding with a \abr{rnn} layer:
\begin{align}
\mathbf{h}_{s_u} &= \rnn{\mathbf{E}_{s_1}, ..., \mathbf{E}_{s_n}}.
\end{align}
Then we fuse the question information into each edge sentence.
An inter attention layer~\cite{seo2016bidirectional} based on the question representation
$\mathbf{h}_q^{(0)}$ first weights each sentence token position $u$
\begin{equation}
  a_{s_u} = \softmax{\mathbf{h}_{s_u} \cdot \mathbf{h}_q^{(0)}};
\end{equation}
the weight is then combined into the question-aware edge sentence $k$'s representation $\mathbf{h}_{s}(k)$:
\begin{equation}
\mathbf{h}_{s}^{(0)}(k) = \sum_u a_{s_u} \mathbf{h}_{s_u}.
\end{equation}
Now that the edges have focused on the evidence that is useful to the question, we
average all edge sentences'
representations $\mathbf{h}_{s}^{k}$ 
into a single edge representation
\begin{align}
    \mathbf{h}_{e}^{(0)} &= \avg{\mathbf{h}_{s}^{(0)}(k)}{k}.
\end{align}

\subsection{Graph Update}
\label{sec:prop}

Given the initial representation of the question~$\mathbf{h}_q^{(0)}$,
nodes~$\mathbf{h}_{v}^{(0)}$, and edges~$\mathbf{h}_{e}^{(0)}$,
\name{}'s \abr{gnn} updates the representations through stacking
multiple layers.  It passes the representations from the question
nodes to the candidate nodes by combining multiple evidence edges'
information.
After this, the representations of the candidate nodes
in the final layer accumulates information from the question nodes
($v_q$), the question text ($q$), the evidence edges ($e_{ij}$), and
previous candidate representations.
These updated node representations are then used to calculate the answer scores.

For each layer $l$, \name{}'s \abr{gnn} first updates the question and
edge representation with a feed forward network (\abr{ffn}) layer
(\textit{Representation Forwarding}), then it scores each edge by its
relevance to the question (\textit{Edge Scoring}), finally passing the
information (\textit{information passing}) from the question entity
nodes (\textit{Question Entity Nodes Update}) to candidate entity
nodes (\textit{Answer Entity Nodes Update}).
We discuss updates going from top to bottom in Figure~\ref{fig:model}.

\paragraph{Question Entity Nodes Update}

\leftnode{}s' representations $\mathbf{h}_{v_q}^{(l)}$ combine the question representation~$\mathbf{h}_q^{(l)}$ and the previous layer's
node representation~$\mathbf{h}_{v_q}^{(l-1)}$, 
\begin{align}
\mathbf{h}_{v_q}^{(l)} &= \ffn{\mathbf{h}_{v_q}^{(l-1)} + \mathbf{h}_q^{(l)}}.
\end{align}

\paragraph{Questions and Edges}

The representations of questions ($q$) and edges ($e$)---initially
represented through their constituent text in their initial
representation---are updated between layer $l-1$ and $l$ through a
feedforward network.
A question's single vector is straightforward,
\begin{align}
  \mathbf{h}_q^{(l)} = & \ffn{\mathbf{h}_q^{(l-1)}},
\end{align}
but edges are slightly more complicated because there may be multiple
sentences linking two entities together.
Thus, each individual sentence $k$ has its representation updated for layer~$l$,
\begin{align}
  \mathbf{h}_s^{(l)}{(k)}= & \ffn{\mathbf{h}_s^{(l-1)}{(k)}},
\end{align}
and those representations are averaged to update the overall edge representation,
\begin{align}
    \mathbf{h}_e^{(l)} = & \avg{\mathbf{h}_s^{(l)}{(k)}}{k}.
\end{align}

\paragraph{Edge Scoring}

Each edge has an edge score~$a_e$; higher edge scores indicate the
edge is more useful to answering the question.
The \abr{gnn} computes an edge score $a_e$ for each edge
representation $\mathbf{h}_e^{(l)}$ based on its similarity to this
layer's question representation:
\begin{align}
    a^{(l)}_e = & \text{Sigmoid}\left(\mathbf{h}_q^{(l)} \cdot \mathbf{h}_e^{(l)}\right).
\label{eq:edge-score}
\end{align}

\paragraph{Information Passing}

The representation of the edge is not directly used to score
\rightnode{}s.
Instead, a representation is created from both the
source \leftnode{}~$\mathbf{h}_{v_q}^{(l)}$ and the combined
edges~$\mathbf{h}_e^{(l)}$ concatenated togther, feeds that through a
feed-forward network (to make the dimension consistent with previous layer and question representation), and weights by the edge score ($a_e$,
Equation~\ref{eq:edge-score}),
\begin{align}
f^{(l)}\left(v_q\xrightarrow{e} v_a\right)&= a_e^{(l)} \ffn{\left[\mathbf{h}_{v_q}^{(l)};\mathbf{h}_e^{(l)}\right]}.
\end{align}

\paragraph{Candidate Entity Nodes Update}

The updated representation for each \rightnode{} 
combines the previous layer's \rightnode{} representation, the question
representation, and the passed information. 
\begin{equation}
  \mathbf{h}_{v_a}^{(l)}= \text{\abr{ffn}}
   \left( \vphantom{\mathbf{h}_{v_a}^{(l-1)}} \right.
  \explain{previous}{\mathbf{h}_{v_a}^{(l-1)}}
                           + \explain{question}{\mathbf{h}_q^{(l)}} +
                           \explain{Information Passing}{\sum_{v_q \in V_q} f^{(l)}(v_q\xrightarrow{e} v_a)}
                              \left. \vphantom{\mathbf{h}_{v_a}^{(l-1)}} \right).
                         \end{equation}
After iterating through several \abr{gnn} layers, the candidate node
representations aggregate information from the question nodes, edges,
and candidate nodes themselves.

\subsection{Answer Scoring}
\label{sec:ans}

Finally, \name{} uses a multi-layer perception (\abr{mlp}) to score the \rightnode{} using the
final layer $L$'s representations,
\begin{equation}
p({v_a; G}) = \mbox{\sigmoid{}}\left(\mbox{\abr{mlp}}(\mathbf{h}^{(L)}_{v_a})\right).
\label{eq:node-update}
\end{equation}
The \abr{gnn} is trained using binary cross entropy
loss over all \rightnode{}s $v_a$.  At test time, it
chooses the answer with the highest $p({v_a, G})$.

%% file: 2020_www_delft/sections/70-experiment.tex
\section{Experiments}
\label{sec:exp}

We evaluate on three datasets with expert-authored (as opposed to crowd-worker) questions.
\qblink~\cite{Elgohary:Zhao:Boyd-Graber-2018} is an entity-centric dataset with human-authored questions. 
The task is to answer the entity the question describes. We use the released dataset for evaluation.
\qanta~\cite{iyyer2014neural} is a \abr{qa} dataset collected from \qb{} competitions. 
Each question is a sequence of sentences providing increased information about the answer entity. 
\triviaqa~\cite{joshi2017triviaqa} includes questions from trivia
games and is a benchmark dataset for \abr{mr}. We use its unfiltered
version, evaluate on its validation set, and split 10\% from its
unfiltered training set for model selection.
Unlike the other datasets, \triviaqa{} is relatively simpler; it
mentions fewer entities per question; as a result \name{} has lower
accuracy.

\input{2020_www_delft/data/stat.tex}

\input{2020_www_delft/data/para.tex}

We focus on questions that are answerable by Wikipedia entities.  To
adapt \triviaqa{} into this factoid setting, we filter out all
questions that do not have Wikipedia title as answer. We keep $~70\%$
of the questions, showing good coverage of Wikipedia Entities
in questions.
All \qblink{} and \qanta{} questions have entities tagged by TagMe.
TagMe finds no entities in 11\% of \triviaqa{} questions; we further
exclude these.
Table~\ref{tab:data} shows the statistics of these three datasets 
sand the fraction of questions with entities.

\subsection{Question Answering Methods}
We compare the following methods:
\begin{itemize*}
\item \quest{}~\cite{Lu:2019:ACQ} is an unsupervised factoid \abr{qa}
  system over text. For fairness, instead of Google results we apply \quest{} on \abr{ir}-retrieved 
Wikipedia documents.

\item \drqa{}~\cite{chen2017reading} is a machine reading model for open-domain 
  \abr{qa} that retrieves documents and extracts answers.
  
\item \docqa{}~\cite{clark2018simple} improves multi-paragraph machine reading, 
and is among the strongest on \triviaqa{}. Their suggested settings 
and pre-trained model on \triviaqa{} are used.

\item \bertet{} fine-tunes \bert{}~\cite{devlin2018bert} on the question-entity name pair to rank candidate entities in \name{}'s  graph.
  
\item \bertsent{} fine-tunes \bert{} on the question-entity gloss sequence pair to rank candidate entities in \name{}'s  graph.
  
\item \memnn{} is a memory network~\cite{weston2014memory} using fine-tuned \bert{}.
It uses the same evidence as \name{} but collapses the 
graph structure (i.e., edge evidence sentences) by concatenating all
evidence sentences into a memory cell.

\item We evaluate our method, \name{}, with \glove{} embedding~\cite{pennington2014glove} and \bert{} embeddings.
\end{itemize*}

\input{2020_www_delft/data/coverage.tex}

\subsection{Implementation}

\input{2020_www_delft/data/overall.tex}

Our implementation uses PyTorch~\cite{paszke2017automatic} and its \abr{dgl} \abr{gnn} library.\footnote{\url{https://github.com/dmlc/dgl}}
We keep top twenty candidate entity nodes in the training and fifty for testing; the top five sentences for each edge is kept.
The parameters of \name{}'s \abr{gnn} layers are listed in Table~\ref{tab:para}. For \name{}-\bert{}, we use \bert{} 
output as contextualized embeddings. 

For \bertet{} and \bertsent{}, we concatenate the question and entity
name (entity gloss for \bertsent{}) as the \bert{} input and
 apply an affine layer and sigmoid activation to the last \bert{} layer of the \abr{[cls]} token; the model
outputs a scalar relevance score.

\memnn{} concatenates all evidence sentences and the node gloss, and combines with the question
as the input of \bert{}.  Like \bertet{} and \bertsent{}, an affine layer and sigmoid activation is
applied on \bert{} output to produces the answer score.

\drqa{} retrieves $10$ documents and then $10$ paragraphs from them; we use the default 
setting for training. During inference, we apply TagMe to 
each retrieved paragraph and 
limit the candidate spans as tagged entities. 
\docqa{} uses the pre-trained model on \triviaqa{}-unfiltered
dataset with default configuration applied to our subset.

%% file: 2020_www_delft/data/stat.tex
\begin{table}[t]
  \centering
  \begin{tabular}{lrrr}
      & \textbf{\qblink{}} & \textbf{\qanta{}} & \textbf{\triviaqa{}}  \\ \toprule
    Training &$42219$ & $31489$ & $41448$\\ 
    Dev &$3276$ & $2211$ & $4620$\\ 
    Test &$5984$ & $4089$ & $5970$\\  \midrule
    \# Tokens  &$31.7 \pm9.4$ & $129.2\pm 32.0$ & $16.5\pm 8.6$ \\
    \# Entities &$6.8\pm2.4$ & $21.2\pm7.3$ & $2.2\pm1.3$\\ \midrule
    \% 1-3 Entities & $9.6\%$& $0$& $86.9\%$\\
    \% 4-6 Entities & $36.7\%$& $0$& $13.1\%$\\
    \% 7-9 Entities & $36.5\%$& $0$& $0$\\
    \% 10+ Entities & $17.1\%$& $100\%$& $0$ \\
  \bottomrule
  \end{tabular}
  \caption{The three expert-authored datasets used for experiments.
    All are rich in entities, but the \qanta{} dataset especially
    frames questions via an answer's relationship with entities
    mentioned in the question.}
  \label{tab:data}
  
\end{table}

%% file: 2020_www_delft/data/para.tex
\begin{table}[t!]
    \centering
    \small
    \begin{tabular}{lp{4cm}}
      \textbf{Layer}   & \textbf{Description} \\ \toprule
        All \abr{rnn} & 1 layer Bi-\abr{gru}, 300 hidden dimension \\
        All \abr{ffn} & 600 dimension, ReLU activation \\
        \abr{mlp} & 2 layers with 600, 300 dimenstions, ReLU activation \\
        Attention & 600 dimension Bilinear  \\
        Self-Attention & 600 dimension Linear \\
        Layers & $L=3$ \\
        \bottomrule
    \end{tabular}
    \caption{Parameters in \name{}'s \abr{gnn}.}
    \label{tab:para}
\end{table}

%% file: 2020_www_delft/data/coverage.tex
\begin{table*}[t]
  \centering
  \begin{tabular}{lrrr}
      & \textbf{\qblink{}} & \textbf{\qanta{}} & \textbf{\triviaqa{}}  \\ \toprule
    \# Candidate Answer Entities per Question &$1607\pm 504$ & $1857 \pm 489$ & $1533\pm 934$\\ 
    Answer Recall in All Candidates
    & $92.4\%$&$92.6\%$  &$91.5\%$\\
    Answer Recall after Filtering
    & $87.6\%$& $83.9\%$ &$86.4\%$ \\ 
    Answer Recall within Two Hops along DBpedia Graph* & 38\% & -- & -- \\
    \midrule
    \# Edges to Correct Answer Node (+) & $5.07\pm2.17$&$12.33\pm5.59$ &$1.87\pm 1.12$\\
    \# Edges to Candidate Entity Node (-) & $2.35\pm0.99$ & $4.41\pm2.02$& $1.21\pm0.35$ \\ 
    \# Evidence Sentences per Edge (+) & $12.3\pm11.1$& $8.83\pm6.17$ & $15.53\pm17.52$ \\ 
    \# Evidence Sentences per Edge (-) &$4.67\pm 3.14$& $4.48\pm1.88$&$3.96\pm3.33$\\
    \bottomrule
  \end{tabular}
  \caption{Coverage and density of generated free-text entity graph. 
  (+) and (-) mark the statistics on correct answer nodes and incorrect nodes, respectively.
  (*) is the result from our manual labeling on 50 \qblink{} questions.
  \label{tab:coverage}}
  
\end{table*}

%% file: 2020_www_delft/data/overall.tex
\begin{table*}[t]
  \centering
  \begin{tabular}{lrrrrr gg rrr}
      & \multicolumn{5}{c}{\textbf{\qblink{}}} &
                                                 \multicolumn{2}{c}{ \textbf{\qanta{}}} & \multicolumn{3}{c}{\textbf{\triviaqa{}}}  \\ 
      & ALL
      & 1-3 & 4-6 & 7-9 & 10+ 
      & ALL
      & 10+
      & ALL
      & 1-3  & 4-6   \\ \toprule
    \quest{} & 0.07 & 0 & 0.09 & 0.09& 0& -&- &- &-&-\\ 
    \drqa{} & 38.3 &  35.4 & 37.5 & 39.2 & 39.6
    &  47.4 & 47.4
    &  40.3 &40.1 & 41.2 \\ 
     \docqa{} & - &- & - &- &-&-
    & - 
    & 49.4 & 49.3 & 49.8 \\ \hline
    \bertet{} & 16.2  & 16.1& 16.3&16.2 &16.4
    &  34.2 & 34.2
     & 25.1 &24.5 & 29.0 \\ 
    \bertsent{} & 34.8  & 34.7& 34.8& 34.7&34.6
    & 54.2 & 54.2
    & 44.5 &44.4 &45.1  \\  
    \memnn{} & 35.8 & 32.7 & 36.1 & 36.5& 34.3
    &  56.1 & 56.1
    & 51.3 & 50.9& 54.0\\
    \midrule
    \name{}-\glove{} & 54.2 & 45.5 & 55.0 & 56.4 & 53.2
    &   65.8 & 65.8
    &  51.3 & 50.1 & 59.5 \\ 
    \name{}-\bert{}  
    & \textbf{55.1} & \textbf{46.8} & \textbf{55.5} & \textbf{57.1} & \textbf{55.5}
    &   \textbf{66.2}  & \textbf{66.2}
    & \textbf{52.0}  &\textbf{ 50.5} & \textbf{61.1} \\
    \bottomrule
  \end{tabular}
  \caption{Answer Accuracy (Exact Match) on ALL questions as well as
    question groups with different numbers of entities: e.g., 0--3 are
    questions with fewer than four entities.  We omit empty ranges
    (such as very long, entity-rich \qb{} questions).  \name{} has
    higher accuracy than baselines, particularly on questions with
    more entities.}

  \label{tab:results}

\end{table*}

%% file: 2020_www_delft/sections/71-evaluation.tex
\section{Evaluation Results}
\label{sec:eva}

Three experiments evaluate \name{}'s graph coverage,
answer accuracy, and source of effectiveness. Then we visualize the
\abr{gnn} attention and examine individual examples.

\begin{figure*}[t]
\centering
\small
 \begin{subfigure}{0.3\textwidth}
        \includegraphics[width=\textwidth]{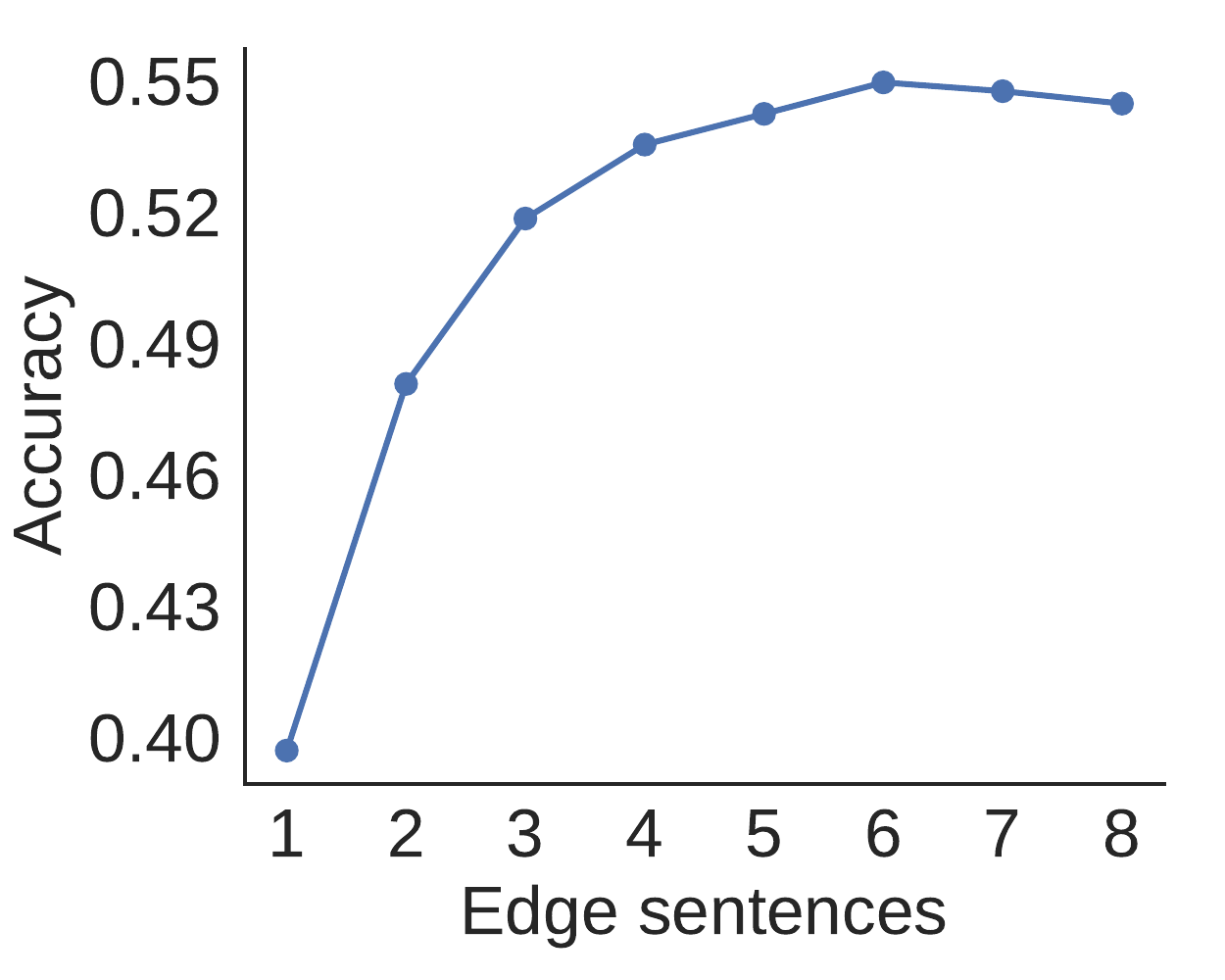}
        \centering
        \caption{Number of Edge Sentences}
    \end{subfigure}
   \begin{subfigure}{0.3\textwidth}
        \includegraphics[width=\textwidth]{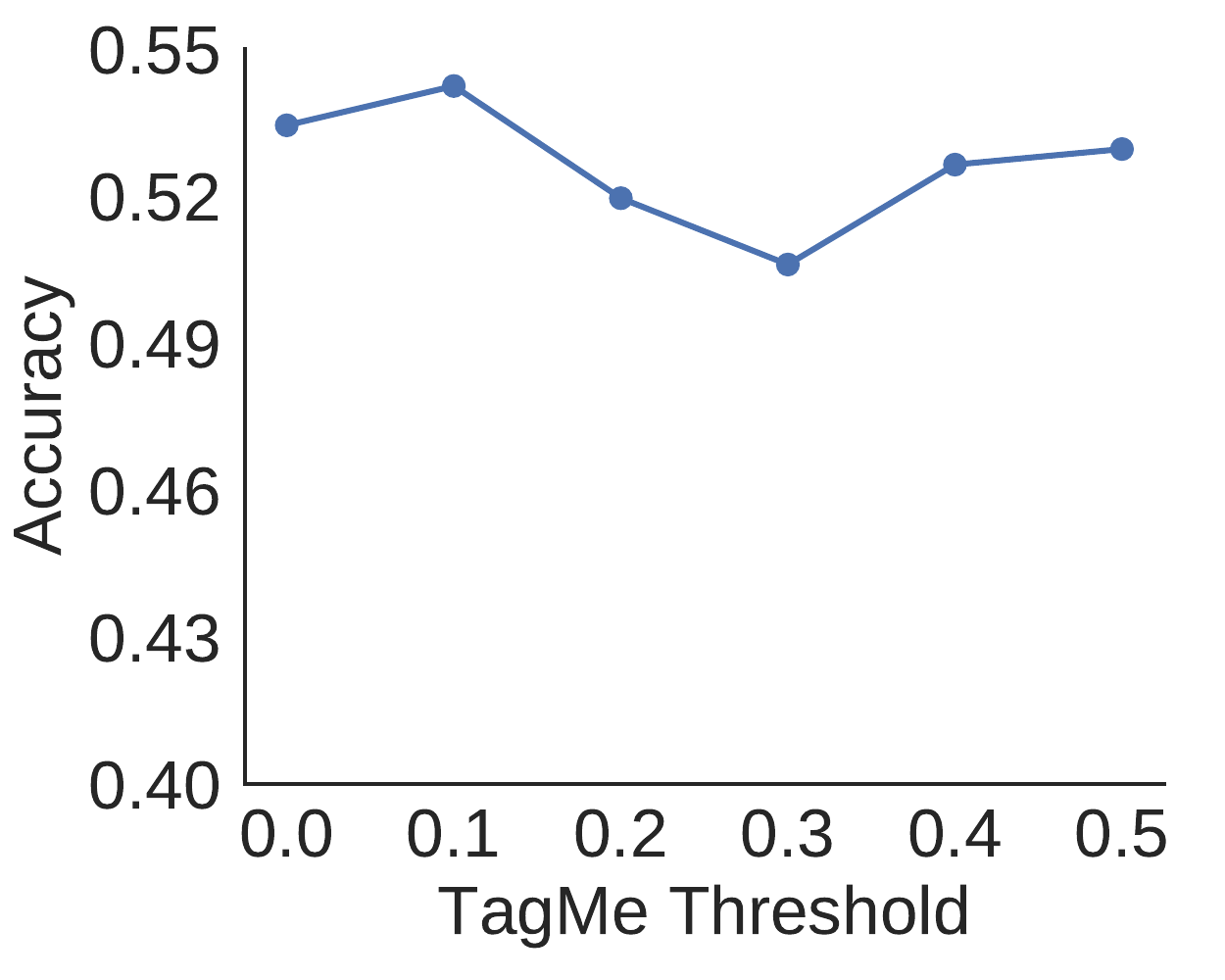}
        \centering
        \caption{TagMe Threshold on Question Entities}
    \end{subfigure}
    \begin{subfigure}{0.3\textwidth}
        \includegraphics[width=\textwidth]{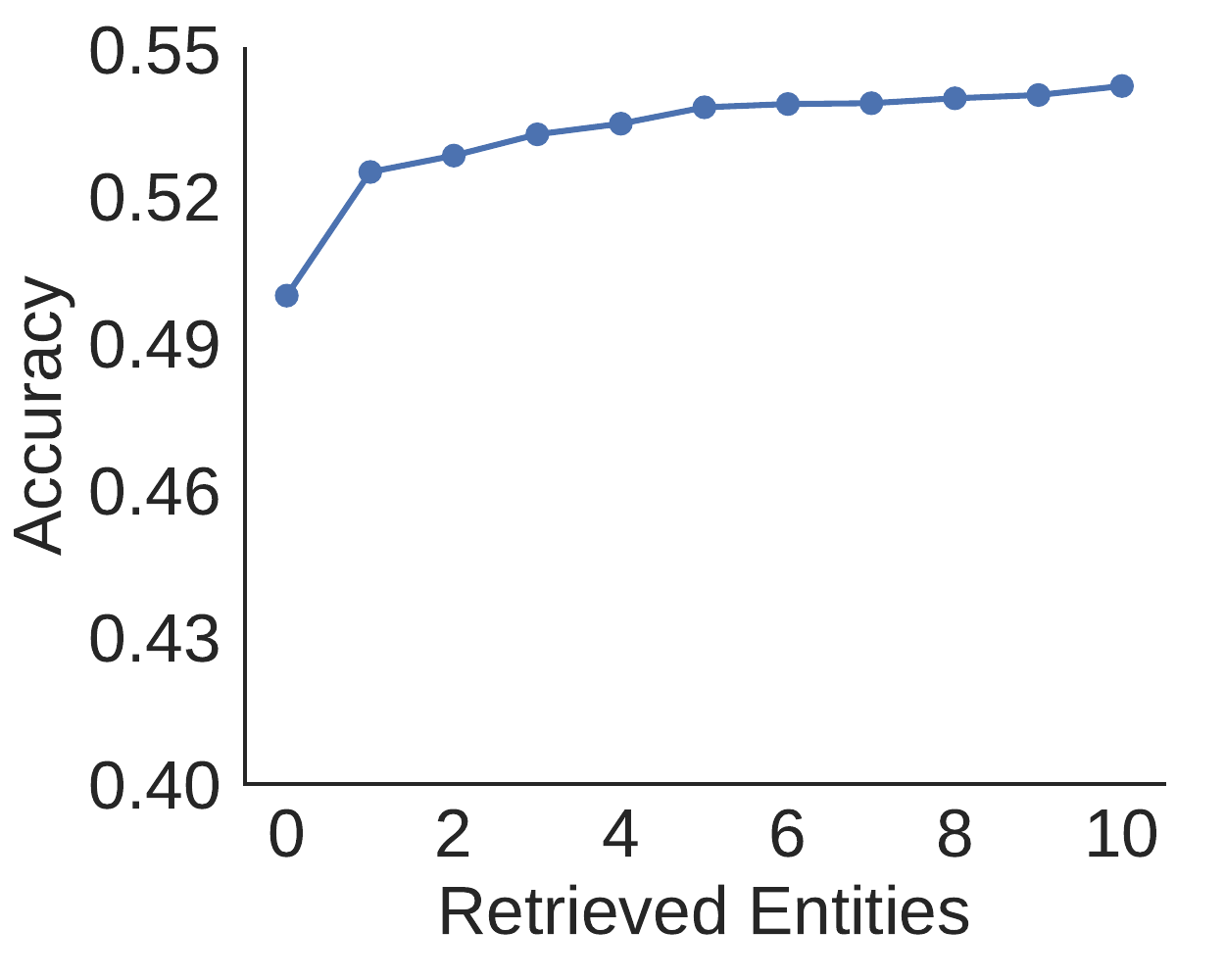}
        \centering
        \caption{Number of Retrieved Entities}
    \end{subfigure}
\caption{Accuracy of \name{} on different variations of Free-Text Knowledge Graphs.
\label{fig:analysis}
}
\end{figure*}

\subsection{Graph Coverage}
\label{sec:coverage}

\name{}'s graph has high coverage (Table~\ref{tab:coverage}).  
Each question is connected to an average of 1500+ candidate nodes; 90\%
of them can be answered by the connected nodes.
After filtering to $50$ candidates, more than 80\% questions are
answerable.
In comparison, we manually examined 50 randomly sampled \qblink{}
questions: only 38\% of them are reachable within two hops in the
DBpedia graph.

\name{}'s graph is dense.
On average there are five (\qblink{}) and twelve (\qanta{}) edges connecting the
correct answer nodes to the question entity nodes.
\triviaqa{} questions have two entities on average and more than one
is connected to the correct answer.  Each edge has eight to fifteen
evidence sentences.

The free-text knowledge graph naturally separates the correct answer by its structure.
Compared to incorrect answers (-), the correct ones (+) are connected by significantly more evidence edges.
The edges also have more evidence sentences.

Aided by free-text evidence, the coverage of the structured graph is
no longer the bottleneck.
The free-text knowledge graph provides enough evidence and frees the
potential of structured ~\abr{qa}.
At the same time, the rich evidence also inevitably introduces noise.
The next experiment examines whether \name{}---given the answer
somewhere in the graph---can find the single correct answer.


\input{2020_www_delft/data/ablation.tex}

\subsection{Answer Accuracy}
\name{} outperforms\footnote{Recall, however, that we exclude
  questions that with no entities or whose answer is not an entity.}
all baselines on both full datasets and dataset subsets based on the
number of entities in a question (Table~\ref{tab:results}).

\quest{} falters on these questions, suggesting that some level of
supervision is required.
On more complicated factoid \abr{qa} datasets \qblink{} and \qanta{},
\name{} improves over \drqa{}, 
the machine reading  baseline.
These datasets require reasoning over multiple sentences (either
within a long question's sentence or across multiple questions); however,
\drqa{} is tuned for single sentence questions.
\name{}---by design---focuses on matching questions' text with
disparate evidence.
In the \abr{mr} benchmark dataset \triviaqa{}, \name{} still beats
\drqa{} (albeit on an entity-focused subset).
It is also better than \docqa{}, one of the strongest models
on \triviaqa{}.
With our Free-Text Knowledge Graph, \name{} better locates 
necessary evidence \emph{sentences} via graph structure, while
\abr{mr} only uses retrieved paragraphs.

\bertet{} fares poorly because it only has entity name information; even
with the help strong pre-trained model, this is 
too limited answer complex questions.
\bertsent{} incorporates the gloss information but lags other methods.
\name{} outperforms both baselines, since it combines useful
text evidence and \abr{kg} connections to answer
the question.

Compared to \memnn{}, which uses the same evidence and \bert{} but
without structure, \name{}'s structured reasoning thrives on complex
questions in \qblink{} and \qanta{}.
On \triviaqa{}, which has fewer than two edges per candidate entity
node, \name{}'s accuracy is close to \memnn{}, as there is not much
structure.

As questions have more entities, \name{}'s relative accuracy
increases. In comparison, almost all other methods' effectiveness
stays flat, even with more evidence from additional question entities.

\subsection{Ablation Study}

We ablate both \name{}'s graph construction and \abr{gnn} components
to see which components are most useful.
We use \qblink{} dataset and \name{}-\glove{} embeddings for these
experiments.
For each ablation, one component is removed while keeping the other
settings constant.

\input{2020_www_delft/data/case.tex}

\paragraph{Graph Ablation} 

The accuracy grows with more sentences per edge until reaching
diminishing returns at six entities (Figure~\ref{fig:analysis}(a)).
Fewer than three sentences significantly decreases
 accuracy, prematurely removing useful information.
It's more effective to leave the \abr{gnn} model to distinguish the
signal from the noise.
We choose five sentences per edge.

Because we retrieve entities automatically (rather than relying on gold
annotations), the threshold of the entity linking process can also be tuned: are more (but noisier) entities better than fewer (but more confident) entities?
Using all tagged entities slightly hurts the result
(Figure~\ref{fig:analysis}(b)), since it brings in uninformative
entities (e.g., linking ``name the first woman in space'' to
``Name'').
Filtering too agressively, however, is also not a good idea, as the
accuracy drops with aggressive thresholds ($>0.2$), removing useful
connections. We choose $0.1$ as threshold.

To see how senstitive \name{} is to automatic entity linking, we
manually annotate twenty questions to see the accuracy with perfect
linking.
The accuracy is on par with Tagme linked questions: both get fifteen
right.
We would need a larger set to more thoroughly examine the role of
linker accuracy.

Recall that \name{} uses not just the entities in the question but
also searches for edges similar to question text.
Figure~\ref{fig:analysis}(c) shows the accuracy when retrieving $n$ additional entities.
The benefits plataeu after three entities.


\begin{figure}[t]
      \begin{center}
      \includegraphics[width=0.9\linewidth]{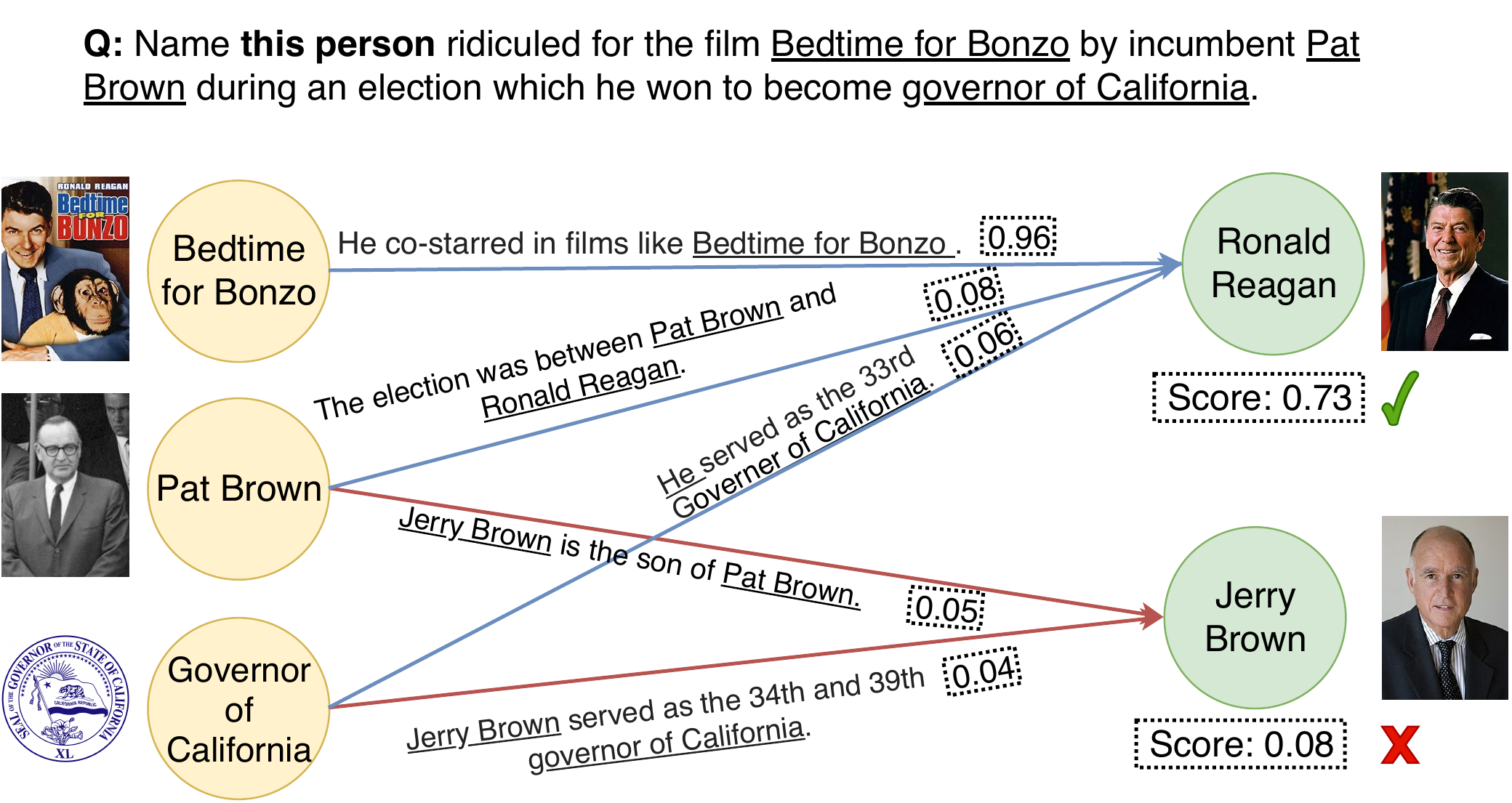}
      \end{center}
      \caption{An example \name{} subgraph. The learned edge weights in \abr{gnn} are in the brackets. }
      \label{fig:graphvis}
\end{figure}

\paragraph{Model Ablation}

In addition to different data sources and preprocessing, \name{}
also has several model components.
We ablate these in Table~\ref{tab:ablation}.
As expected, both node (gloss) and edge evidence 
help; each contributes $\sim 10\%$ accuracy.
Edge importance scoring, which controls the weights of the information
flow from \leftnode{} to \rightnode{}s, provides $\sim 3\%$
accuracy.
The input representation is important as well; the self-attention
layer contributes $\sim 2.2\%$ accuracy.

\subsection{Graph Visualization}

Figure~\ref{fig:graphvis} shows a question with \abr{gnn} output:
the correct \rightnode{} \candidate{Ronald Reagan}
connects to all three \leftnode{}s.
The edge from \question{Bedtime for Benzo} to \candidate{Reagan} is
informative---other \rightnode{}s (e.g., politicians like
\candidate{Jerry Brown}) lack ties to this cinema masterpiece.
The \abr{gnn} model correctly (weight $0.96$) favors this edge.
The other edges are less distinguishable.
For example, the edge from \question{Governor of California} to
\candidate{Ronald Reagan} and \candidate{Jerry Brown} are both
relevant (both were governors) but unhelpful.
Thus, the \abr{gnn} has similar weights ($0.06$ and $0.04$) for
both edges, far less than \question{Bedtime for Bonzo}.
By aggregating edges, \name{}'s \abr{gnn} selects the correct
answer.

\input{2020_www_delft/sections/72-error.tex}

%% file: 2020_www_delft/data/ablation.tex
\begin{table}[t]
  \centering
  \small
  \begin{tabular}{lrr}
    \textbf{Ablation}  & \multicolumn{2}{r}{\textbf{Accuracy}} \\ \toprule
    No Node Pruning & 42.6 & -21.4\% \\ 
    No Gloss Representation & 49.2 & -9.3\% \\ 
    No Edge Evidence Sentence & 48.8 & -10.0\% \\ 
    No Edge Importance & 52.6 & -3.0\% \\ 
    No Self Attention Layer & 53.0 & -2.2\% \\ 
    \name{}-\glove{} &54.2 & -- \\
    \bottomrule
  \end{tabular}
  \caption{Ablation Study of \name{}-\abr{g}{\small love} on \qblink{} (ALL questions). 
    Each \name{} variant removes one component and keeps everything else fixed. }
    \label{tab:ablation} 
\end{table}

%% file: 2020_www_delft/data/case.tex
\begin{table*}[t]
  \centering
  \small
  \begin{tabular}{p{0.5cm}p{7.5cm}p{8cm}}
      id & Example  &Explanation  \\ \toprule
      1(+)&\textbf{Q}: This general was the leader of the nationalist side in the 
      \question{civil war} and ruled until his death in 1975. 
      He kept \question{Spain} neutral in the \question{Second World War} and is still dead.

      \textbf{A}: Francisco Franco  \hphantom{\dots\dots} \textbf{P}: \candidate{Francisco Franco}
      
      & The question directly maps to evidence sentence ``After the nationalist victory in the Spanish Civil War, until his (\candidate{Francisco Franco}) death in 1975''.\\ 

      2(+)&\textbf{Q}: This \question{New England Patriots} quarterback was named \question{Super Bowl MVP}. He had three touchdown passes during the game: one each to \question{Deion Branch}, \question{David Givens}, and \question{Mike Vrabel}.

      \textbf{A}: Tom Brady \hphantom{\dots\dots} \textbf{P}: \candidate{Tom Brady}

      & Substantial evidence points to \candidate{Tom Brady} (correct): ``\candidate{Tom Brady}  plays for \question{New England Patriots}'', and ``\candidate{Tom Brady} had touchdown passes with \question{Deion Branch}''.
      \name{} aggregates evidence and makes the correct prediction. Without the graph structure, \memnn{} instead 
      predicts \candidate{Stephon Gilmore} (Another \question{New England Patriot}{}).\\
      
      3(-)&\textbf{Q}: Name this \question{European} nation which was divided into Eastern and Western regions after \question{World War II}. 

      \textbf{A}: Germany  \hphantom{\dots\dots}  \textbf{P}: \candidate{Yumen Pass}

      & No informative question entities that would lead to the key evidence sentence ``Germany divides into East Germany and West Germany''.\\ 

      4(-)&\textbf{Q}: \question{Telemachus} is the son of this hero, who makes a really long journey back home after the \question{Trojan War} in an epic poem by \question{Homer}.

      \textbf{A}: Odysseus  \hphantom{\dots\dots}  \textbf{P}: \candidate{Penelope}

      & \name{} can't make right prediction since the wrong candidate \candidate{Penelope} (Odysseus's wife) shares 
      most of the extracted evidence sentences with the correct answer (e.g., their son \question{Telemachus}). 
      \\\bottomrule

  \end{tabular}
  \caption{Four examples from \qblink{} dataset with \name{} output. Each example has a question (Q), gold answer (A) and \name{} predcition (P), along with an explanation of what happened. The first two are correct (+), while the last two are wrong (-). 
  }
  \label{tab:example}
\end{table*}

%% file: 2020_www_delft/sections/72-error.tex
\subsection{Case Study}

To gain more insights into \name{} model's behavior, we further sample
some examples from \qblink{}.  Table~\ref{tab:example} shows two
positive examples ($1$ and $2$) and two negative examples ($3$ and
$4$).  \name{} succeeds on these cases: with direct evidence
sentences, \name{} finds the correct answer with high confidence
(Example 1) or with multiple pieces of evidence, \name{} could
aggregate different pieces together, and make more accurate
predictions (Example 2).  However some common sources of error
include: too few informative entities in the question (Example 3) or
evidence that overlaps too much between two \rightnode{}s.


%% file: 2020_www_delft/sections/60-related.tex
\section{Related Work: Knowledge Representation for QA}
\label{sec:bg}

\name{} is impossible without the insights of 
traditional knowledge bases for question answering and question
answering from natural language, which we combine using graph neural
networks.
This section describes how we build on these subfields.

\subsection{Knowledge Graph Question Answering}

With knowledge graphs (\abr{kg}) like
Freebase~\cite{bollacker2008freebase} and
DBpedia~\cite{mendes2011dbpedia} enable question answering using their
rich, dependable structure.
This has spurred \abr{kg}-specific \abr{qa} datasets on general domain
large scale knowledge graphs: WebQuestions~\cite{berant2013semantic},
SimpleQuestions~\cite{bordes2015large}, and special-domain \abr{kg}s,
such as WikiMovies~\cite{miller2016key}.
In turn, these new datasets have prompted special-purpose 
\abr{kgqa} algorithms.
Some convert questions to semantic parsing problems and execute the
logical forms on the graph~\cite{cai2013large, kwiatkowski2013scaling,
  reddy2014large, yih2015semantic}.
Others use information extraction
 to first extract question related information in \abr{kg}
and then find the answer~\cite{bao2014knowledge, yao2014information,
  gardner2017open}.

These work well on questions tailored for the underlying \abr{kg}.
For example,
WebQuestions guarantee its questions can be answered by
Freebase~\cite{berant2013semantic}.
Though modern knowledge graphs have good coverage on
entities~\cite{cxthesis}, adding relations takes time and
money~\cite{Paulheim2018HowMI}, often requiring human
effort~\cite{bollacker2008freebase} or scraping human-edited
structured resources~\cite{mendes2011dbpedia}.
These lacun\ae represent impede broader use and
adoption.

Like \name{}, \abr{quest}~\cite{Lu:2019:ACQ} seeks to address this by
building a noisy quasi-\abr{kg} with nodes and edges, consisting of
dynamically retrieved entity names and relational phrases from raw
text.
Unlike \name{}, this graph is built using existing Open Information
Extraction (\abr{ie}).
Then it answers questions on the extracted graph.
Unlike \name{}, which is geared toward recall, \abr{ie} errs toward
precision and require regular, clean text.
In contrast, many real-world factoid questions contain linguistically
rich structures, making relation extraction challenging.
We instead directly extract free-text sentences as indirect relations
between entities, which ensures high coverage of evidence information
to the question.

Similarly, {\abr{graft}-\abr{n}\small et}~\cite{sun2018open} extends an existing \abr{kg} with 
text information.
It grafts text evidence onto \abr{kg} nodes but retains the original \abr{kg} relations.
It then reasons over this graph to answer \abr{kg}-specific questions.
\name{}, in contrast, grafts text evidence onto both nodes and edges
to enrich the relationships between nodes, building on
the success of unconstrained ``machine reading'' \abr{qa} systems.

\subsection{Question Answering over Text}

Compared with highly structured \abr{kg}, unstructured text
collections (e.g., Wikipedia, newswire, or Web scrapes) is cheaper but
noisier for \abr{qa}~\cite{chen2018neural}.
Recent datasets such as \squad{}~\cite{rajpurkar2016squad},
\triviaqa{}~\cite{joshi2017triviaqa}, \abr{ms marco}~\cite{marco} and
natural questions~\cite{kwiatkowski2019natural} are typically solved
via a coarse search for a passage (if the passage isn't given) and
then finding a fine-grained span to answer the question.

A rich vein of neural readers match the questions to the given
passages and extract answer spans from them~\cite[inter
  alia]{seo2016bidirectional, yu2018qanet}.  Its popular solutions
include \abr{bidaf}, which matches the question and document passages
by bi-directional attention flows~\cite{seo2016bidirectional},
\abr{qan}{\small et}, which enriches the local contexts with global
self-attention~\cite{yu2018qanet}, and pre-training methods such as
\abr{bert}~\cite{devlin2018bert} and \abr{xln}{\small
  et}~\cite{yang2019xlnet}.

The most realistic models are those that also search for a passage:
\drqa{}~\cite{chen2017reading} retrieves documents from Wikipedia and
use \abr{mr} to predict the top span as the answer, and
\abr{orca}~\cite{lee-19} trains the retriever via an inverse cloze
task.
Nonetheless, questions mainly answerable by \drqa{} and \abr{orca} only require
single evidence from the candidate sets~\cite{min2018efficient}.
\name{} in contrast searches for edge evidence and nodes that can
answer the question; this subgraph often corresponds to the same
documents found by machine reading models.
Ideally, it would help synthesize information across \emph{multiple}
passages.

Multi-hop \abr{qa}, where answers require require assembling
information~\citep{welbl2018constructing, yang2018hotpotqa}, is a
\emph{task} to test whether machine reading systems can
synthesize information.
\abr{h}{\small otpot}\abr{qa}~\cite{yang2018hotpotqa} is the 
 multi-hop \abr{qa} benchmark: each answer is a text span requiring 
one or two hops.
Several models~\cite{qiu-etal-2019-dynamically,
  ding-etal-2019-cognitive, min2019multi} solve
this problem using multiple \abr{mr} models to extract multi-hop
evidence.
While we focus on datasets with Wikipedia entities as
answers, expanding \name{} to span-based answers (like \abr{h}{\small otpot}\abr{qa}) is a
natural future direction.

\subsection{Graph Networks for \abr{qa}}

\name{} is not the first to use graph neural networks~\cite[inter
  alia]{scarselli2009graph, kipf2016semi, schlichtkrull2017modeling}
for question answering.
\abr{e}{\small ntity}-\abr{gcn}~\cite{de2018question},
\abr{dfgn}~\cite{qiu-etal-2019-dynamically}, and
\abr{hde}~\cite{tu-etal-2019-multi} build the entity graph with entity
co-reference and co-occurrence in documents and apply \abr{gnn} to
the graph to rank the top entity as the answer.
\abr{c}{\small og}\abr{qa}~\cite{ding-etal-2019-cognitive} builds the
graph starting with entities from the question, then expanding the
graph using extracted spans from multiple \abr{mr} models as candidate
span nodes and adopt \abr{gnn} over the graph to predict the answer
from span nodes.
All these methods' edges are co-reference between entities or binary
scores about their co-occurrence in documents; \name{}'s primary
distinction is using free-text as graph edges which we then represent
and aggregate via a \abr{gnn}.

Other methods have learned representations of relationships between entities.
\abr{nubbi}~\cite{chang-09c} used an admixture over relationship
prototypes, while Iyyer et al.~\cite{iyyer-16} used neural dictionary
learning for analyzing literature.
%
\name{} draws on these ideas to find similarities between 
passages and questions.

%% file: 2020_www_delft/sections/80-conclusions.tex
\section{The View Beyond \name{}}
\label{sec:conclu}

Real-world factoid \abr{qa} requires answering diverse questions across domains.
Relying on existing knowledge graph relations to answer these questions often leads to highly accurate but brittle systems:
they suffer from low coverage.
To overcome the bottleneck of structure sparsity in existing knowledge graphs,
\name{} inherits 
\abr{kgqa}-style reasoning with the widely available free-text evidence.
\name{} builds a high coverage and dense free-text knowledge graph, using
natural language sentences as edges.
To answer questions, \name{} grounds the question into the related
subgraph connecting entities with free-text graph edges and then uses
a graph neural network to represent, reason, and select the answer
using evidence from both free-text and the graph structure.

Combining natural language and knowledge-rich graphs is a common
problem:
e-mail and contact lists,
semantic ontologies and sense disambiguation,
and semantic parsing.
Future work should explore whether these approachs are also useful for
dialog, language modeling, or \textit{ad hoc} search.

More directly for question answering, more fine-grained reasoning
could help solve the example of Table~\ref{tab:example}:
while both Odysseus and Penelope have Telemachus as a son, only
Odysseus made a long journey and should thus be the answer.
Recognizing specific properties of nodes either in a traditional
\abr{kg} or in free text could resolve these issues.